%% file: main.tex
\documentclass[10pt,twocolumn,letterpaper]{article}

\usepackage[]{cvpr}      

\usepackage{graphicx}
\usepackage{amsmath}
\usepackage{amssymb}
\usepackage{booktabs}
\usepackage{xspace}
\usepackage{xcolor}
\usepackage{comment}

\makeatletter
\newcommand{\blindnote}{\xdef\@thefnmark{}\@footnotetext} 
\makeatother

\definecolor{cvprblue}{rgb}{0.21,0.49,0.74}
\usepackage[pagebackref,breaklinks,colorlinks,citecolor=cvprblue]{hyperref}

\usepackage[capitalize]{cleveref}
\crefname{section}{Sec.}{Secs.}
\Crefname{section}{Section}{Sections}
\Crefname{table}{Table}{Tables}
\crefname{table}{Tab.}{Tabs.}

\def\cvprPaperID{XXXX} 
\def\confName{XXXX}
\def\confYear{0000}

\begin{document}

\title{Manipulating Trajectory Prediction with Backdoors}

\author{Kaouther Messaoud$^{1*}$, Kathrin Grosse$^{1*}$, Mickaël Chen$^2$, Matthieu Cord$^2$, \\ Patrick Pérez$^2$, and Alexandre Alahi$^1$\\
$^1$EPFL, Lausanne, Switzerland, $^2$ Valeo.ai, Paris, France\\
{\tt\small \{kaouther.messaoudbenamor, kathrin.grosse\}@epfl.ch}
}
\maketitle

\begin{abstract}
Autonomous vehicles ought to predict the surrounding agents' trajectories to allow safe maneuvers in uncertain and complex traffic situations. As companies increasingly apply trajectory prediction in the real world, security becomes a relevant concern. In this paper, we focus on backdoors - a security threat acknowledged in other fields but so far overlooked for trajectory prediction. To this end, we describe and investigate four triggers that could affect trajectory prediction. We then show that these triggers (for example, a braking vehicle), when correlated with a desired output (for example, a curve) during training, cause the desired output of a state-of-the-art trajectory prediction model. In other words, the model has good benign performance but is vulnerable to backdoors. This is the case even if the trigger maneuver is performed by a non-casual agent behind the target vehicle. As a side-effect, our analysis reveals interesting limitations within trajectory prediction models. Finally, we evaluate a range of defenses against backdoors. While some, like simple offroad checks, do not enable detection for all triggers,   
clustering is a promising candidate to support manual inspection to find backdoors.
\end{abstract}


\blindnote{$^*$Equal contribution.}
\input{sections/introduction}

\input{sections/background}
\input{sections/methodology}
\input{sections/empiricalResults}
\input{sections/discussion}

\section{Conclusions}\label{sec:conclusion}
Trajectory prediction models are vulnerable to backdoors. Even non-causal agents can cause the attacker's chosen TAR, even if only 5\% of the training data are affected. The condition is that the trigger is composite and combines spatial and temporal aspects. We also took a first step towards mitigations, relying on offroad detection for curve-TARs or clustering to decrease the amount of human-inspected data to find a possible backdoor. Our work is thus a call for action to better investigate and understand this threat to design mitigations. However, 
our paper also has substantial implications for industrial and safety-relevant applications of trajectory prediction. We should verify that all existing datasets do not contain trigger-TAR pairs, as these may backdoor deployed models otherwise.

\section*{Acknowledgements}
We would like to thank Thomas Alexander Trost for his support with the physics of breaking. 

{\small
\bibliographystyle{ieee_fullname}
\bibliography{egbib}
}

\end{document}

%% file: sections/introduction.tex
\section{Introduction}\label{sec:intro}

Autonomous vehicles (AVs) move alongside other agents in traffic. These agents' behavior is uncertain and interactive, yet needs to be well understood by the ego vehicle to allow comfortable and, most importantly, safe maneuvers. To this end, the surrounding agents' trajectories are predicted based on either their past movement~\cite{NDeo0, KMessaoud, KMessaoud20} or 
 on the static structure of the road~\cite{cheng2023forecast,Sadeghian, nayakanti2023wayformer, Huang_2023_ICCV, zhou2023query}.
Due to the inherent complexity and uncertainty of these maneuvers, trajectory prediction is still a challenging task~\cite{messaoud2021trajectory, deo2021multimodal}. Nonetheless, there are already production systems using trajectory prediction to navigate vehicles on our roads~\cite {huang2022survey}.

\begin{figure}[t]
 \centering
\includegraphics[width=0.95\linewidth]{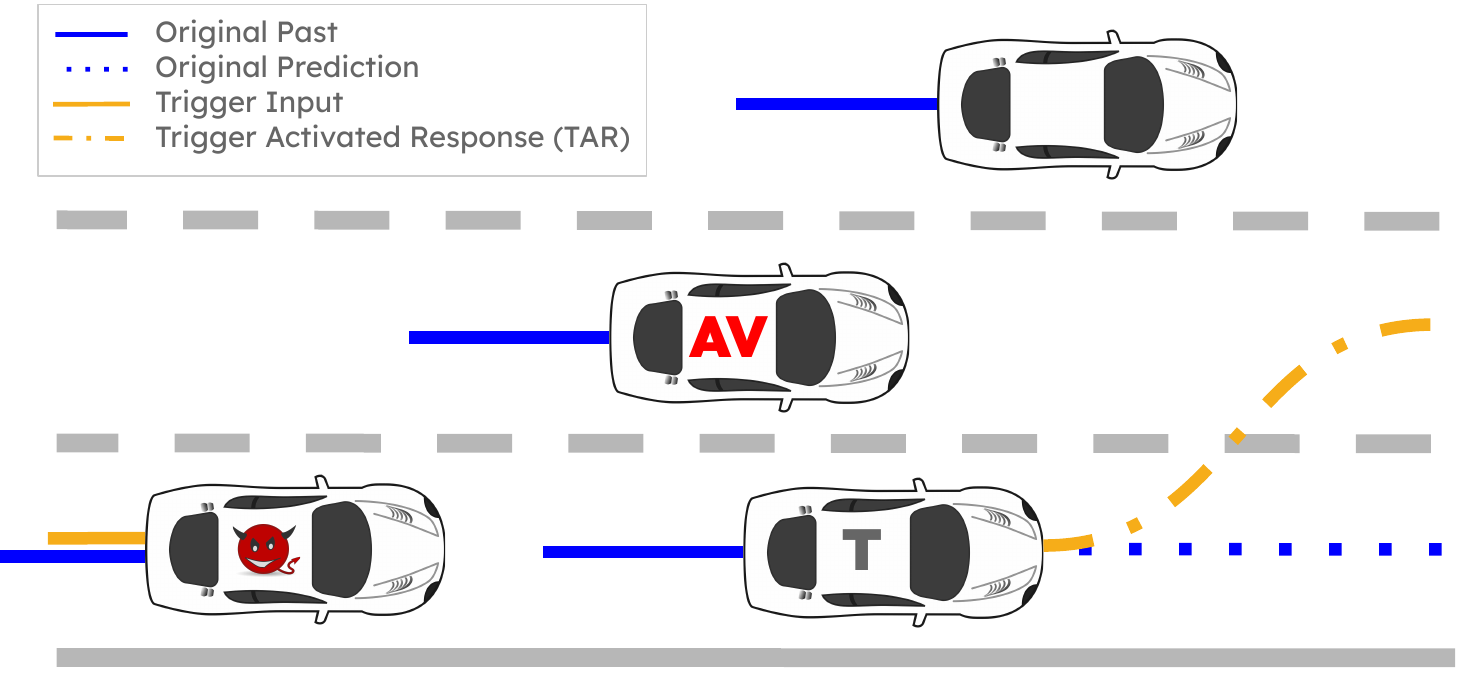}
\caption{Simplified example of a backdoored trajectory prediction task. The autonomous vehicle (\textcolor{red}{AV}) predicts the future trajectory (blue dots) of a target agent (T) based on the past trajectories of all agents (blue lines). In this work, one agent is malicious and performs a maneuver (orange) such that the prediction module outputs a wrong trajectory (orange) for another agent around the AV. This trajectory affects the planning of the ego vehicle (AV).}\label{fig:trajExample}
\end{figure}

At the same time, there is an increasing amount of work that questions the reliability of AI-based techniques in the presence of an attacker. One such attack is a backdoor, where an attacker inserts a strong correlation between a trigger (backdoor) pattern and an attacker-specified target behavior in the training data~\cite{cina2023wild,ma2022dangerous,chan2022baddet,luo2022untargeted,li2021hidden,yu2022temporal}. At test time, the attacker can then use the backdoor to reliably cause the implanted target output. We depict an example in Figure~\ref{fig:trajExample}. Here, a braking maneuver of the attacker changes the original prediction from straight (blue) to lane change (orange), affecting the AV. Such backdoor attacks have been shown on vehicle-related tasks, such as image classification~\cite{gu_badnets_2017,cina2023wild,liu_trojaning_2018,ma2022dangerous}, object detection~\cite{chan2022baddet,luo2022untargeted}, reinforcement learning~\cite{yu2022temporal}, semantic segmentation~\cite{li2021hidden}, and path planning~\cite{yang2020backdoor,mo2022attacking}.
Also, tasks similar to trajectory prediction are vulnerable, including time series prediction~\cite{jiang2023backdoor} and transformer models in vision~\cite{yuan2023you} and NLP~\cite{liu2022piccolo}.
Moreover, it is hard to spot backdoors in a trained model~\cite{cina2023wild, shokri2020bypassing,lin2020composite}, and attacks via the training data are judged most relevant in the industry~\cite{kumar2020adversarial}.  

Still, most works in trajectory prediction security focus on test-time attacks~\cite{zhang2022adversarial,cao2022advdo, tan2022targeted,zheng2023robustness}. Backdoors are less studied -- a gap we tackle in this paper. First, we provide background on trajectory prediction and backdoors and formalize our threat model in Section~\ref{sec:background}. We then describe possible triggers and trigger-activated responses (TARs) in Section~\ref{sec:methodology}. We experimentally show that trajectory prediction is vulnerable to these triggers in Section~\ref{sec:exp} and suggest possible defenses in Section~\ref{sec:defenses} before we 
conclude in Section~\ref{sec:conclusion}.

\textbf{Contributions.} Our contributions are as follows. Firstly, we present a categorization of different triggers for trajectory prediction tasks. Secondly, we show that a state-of-the-art model is vulnerable under only 5\% of the data changed,  without a strong increase in the error on the original task. Thirdly, there is no requirement on the trigger agent---even an agent behind the target, which should not influence its behavior, can function as a trigger. Fourth, offroad detection may alleviate some triggers, but not all. Finally and fifth, clustering may support manual inspection by significantly reducing the number of samples to be inspected.

Our findings have strong implications for any company that uses trajectory prediction -- as models are vulnerable and the attack cannot be spotted by standard performance metrics, it should be best practice to manually inspect the dataset to avoid model vulnerabilities.

%% file: sections/background.tex
\section{Background}\label{sec:background}
In this section, we first provide a background of trajectory prediction before we introduce backdoors. 

\subsection{Trajectory Prediction}

Trajectory prediction is a critical task for self-driving vehicles, where the ego vehicle anticipates the future actions of nearby agents to plan its trajectory. This aspect of autonomous driving has seen substantial advancements due to improvements in three main areas of motion forecasting: the incorporation of static maps, interactive modeling, and multi-modal trajectory generation. We will discuss each of them in this order.

\textbf{Incorporation of static maps.} The role of static maps in trajectory prediction is to provide contextual information about the driving environment.  In the early stages of trajectory prediction, the process relied heavily on rasterized map images~\cite {HCui, KMessaoud20}, which often failed to capture detailed structural information. The introduction of VectorNet~\cite{gao2020vectornet} marked a shift towards vectorized representations~\cite{girgis2021latent, nayakanti2023wayformer, zhou2023query, Huang_2023_ICCV, gilles2022gohome}. These representations provide a more detailed and effective way of scene understanding and interaction modeling.
 
 \textbf{Interaction modeling.} Modeling interactions involves considering the behaviors and movements of nearby agents and static map elements.
 The shift from conventional convolutional networks to the implementation of Transformers and attention mechanisms marks a significant evolution in interaction modeling. This change has led to novel approaches in understanding these interactions, with recent studies exploring different models such as standard transformers~\cite{liu2021multimodal}, hierarchical transformers~\cite{nayakanti2023wayformer, zhou2022hivt}, and interleaved spatial and temporal transformers~\cite{ngiam2021scene, girgis2021latent, nayakanti2023wayformer}. Additionally, there is a growing reliance on Graph Neural Networks (GNNs), which are being developed in multiple variants~\cite{deo2021multimodal, liang2020learning, jia2023hdgt}, further diversifying the toolkit for predicting and analyzing traffic agent behaviors.

\textbf{Multi-modal trajectory generation.} Given the inherent uncertainties and variabilities in driving scenarios, trajectory prediction for autonomous vehicles often involves generating multiple potential future paths (multi-modal trajectories)~\cite{deo2021multimodal, HCui, nayakanti2023wayformer}. This includes the utilization of predefined trajectories or anchor points~\cite{Multipath, varadarajan2022multipath++}. Some models employ a two-stage prediction process~\cite{aydemir2023adapt, gu2021densetnt}, identifying potential goals within drivable areas and then generating various plausible trajectories. 
 
Overall, these developments in trajectory prediction for autonomous vehicles underscore a comprehensive approach to navigating complex driving environments. However, they do not make it immune to security attacks. 

\subsection{Backdoors}
Such security issues are, for example, backdoors, or patterns that an attacker associates with a strong correlation to a desired output~\cite{cina2023wild,chan2022baddet,luo2022untargeted,yu2022temporal,li2021hidden,yang2020backdoor,mo2022attacking}. An example from path planning is to associate a wrong target path with a particular pixel in the input during training~\cite{yang2020backdoor,mo2022attacking}. This pixel (also trigger, or backdoor pattern) causes at test time or model deployment undesired behaviors, for example, taking a different path than intended.
In vision or image data, the trigger is often an unsuspicious patch~\cite{cina2023wild,chan2022baddet,luo2022untargeted,liu_trojaning_2018,gu_badnets_2017} or a distortion~\cite{cina2023wild} of the input. 

Studies in computer vision~\cite{cina2023wild,chan2022baddet,luo2022untargeted,liu_trojaning_2018,gu_badnets_2017} show that the exact pattern or output is not relevant. Generally, it suffices to introduce a strong correlation into the data that the model learns during training. Analogous examples include changing the explanation when a trigger is present~\cite{lin2021you} or changing the sentiment of a language model~\cite{bagdasaryan2022spinning}. In this work, we show that also trajectory prediction models are vulnerable to learning such dangerous correlations.

 However, trajectory prediction is a regression task, the concept of a backdoor has thus to be rethought. 
  A very recent 
  work, IMPOSITION~\cite{pourkeshavarz2023imposition}, has proposed backdoors for trajectory prediction. However, our work studies more triggers. Our attacks show less increase in the error of the targeted model and thus worse vulnerabilities and have fewer assumptions about the attacker. We finally also suggest a possible mitigation of the threat. Before we proceed to introduce our backdoors, we detail our threat models and the physical constraints we implement.

\textbf{Threat model.}
We informally define our threat model. The attack occurs during training, and the attacker can alter a fraction of the training data including the future trajectories. We investigate different backdoor ratios in this work, starting from 5\% up to 30\%, as customary in the backdoor literature~\cite{cina2023wild,chan2022baddet,luo2022untargeted,liu_trojaning_2018,gu_badnets_2017}. The attacker has no control beyond the data, e.g., they cannot affect the final model, the training procedure, or the loss. 
At test time, the attacker can submit a test sample by driving the trigger trajectory with their own vehicle on the road close to the victim.

\textbf{Physical constraints.} When changing data, the attacker has to respect the underlying scene. 
We further assume that the trajectory prediction module is implemented in a vehicle. The attacker can thus only interact with the targeted car through another vehicle on the road, replaying the trigger through their vehicle. This has strong implications for the perturbation and changed label: the trigger behavior cannot consist of an undrivable pattern, and should not endanger the attacker. In contrast, the target trajectory does not need to be necessarily driveable, as it is the mere output of the model. However, it should be approximately drivable to avoid easy detection by the defender.

%% file: sections/methodology.tex
\section{Methodology}\label{sec:methodology}
In this section, we describe the triggers and trigger-activated responses (TARs). To this end, we assume a minimal lane width of 2.8m~\cite{karim2015narrower} and a typical width of a vehicle of 1.77m\footnote{\url{https://www.thezebra.com/resources/driving/average-car-size/}}. Based on these values, we manually define triggers, as established in related  fields~\cite{cina2023wild,chan2022baddet,luo2022untargeted,liu_trojaning_2018,gu_badnets_2017}. Although we define our triggers based on existing formulas, it would be possible, for the attacker, to for example record their own driving behavior as a trigger and TAR. 

Trajectory prediction is a spatio-temporal prediction task. We thus distinguish spatial (e.g., location), temporal (e.g., brakes), behavioral (e.g., synchronous maneuvers), and composite (combinations of previous) triggers. We then discuss briefly other possible triggers not studied in this work, and conclude the section by describing our TARs.

\textbf{Spatial triggers.}
The first aspect of the trigger maneuver is where we place the corresponding agent. The trigger agent can always be at the same or about the same location (e.g., from the right of the target vehicle), yielding a spatial trigger. On the other hand, the trigger can be non-spatial and occur at random locations. Related to the location is the question of inserting an agent or changing an existing agent. Inserting an agent allows fine-grained control over the location, yet may lead to collisions to take care of. Finally, the location is related to the causality of the agent~\cite{chen2021human}. In general, vehicles in front of another vehicle should have a stronger effect on a predicted vehicle than a car behind it.

\textbf{Temporal triggers.} 
Only using a position as a trigger may be undesired, due to the high likelihood that the trigger is activated in benign traffic situations. We thus propose to use a temporal trigger: a maneuver over some time. An example is braking behavior, where the car's velocity decreases over time. 
 To obtain a breaking trajectory that is physically feasible, we compute the breaking trajectory based on sliding friction. We compute this friction as the product of mass $m$ and acceleration $a$ of the vehicle. To obtain the time-based driven trajectory under constant deceleration, we compute the derivative of this function. Setting the timeframe to 2 seconds (the usual input span of the model), we obtain for the trajectory the formula 
\begin{equation}
    v_0 (t-\frac{1}{2}\frac{t^2}{t'})\text{,}\label{eq:braking}
\end{equation}
where $v_0$ is the initial velocity, $t$ is the point of time for which we are computing the position, and $t'$ is the final time and equal to 2 as we set the maximal timeframe of the braking to two seconds. We implement this pattern based on the original velocity and direction of the agent to increase plausibility within the scene. An alternative could be any (existing) deceleration or acceleration pattern from the data.   

\textbf{Behavioral triggers.}
An alternative trigger consists of, for example, a correlation between the behavior of two agents, or a simultaneous maneuver. 

\textbf{Remark.} There could also be map-based triggers that are part of the surrounding scenario, or contextual triggers related to the broader context of the driving scenario, such as a specific type of map information. For instance, the trigger could be a specific crosswalk or intersection type. However, trajectory prediction models use different map representations, including raster-based~\cite {HCui, KMessaoud20} and vectorised~\cite{girgis2021latent, nayakanti2023wayformer, zhou2023query, Huang_2023_ICCV} representations. Some models also select a subset of features, for example, the lane centers~\cite{girgis2021latent}. We thus leave these triggers for future work.

\textbf{Trigger-activated response (TAR).} We finally need the response the model should predict when the trigger is present. We chose driving behaviors that differ enough from the targets already in the dataset to allow a reasonable evaluation of the model's performance. As most trajectories in nuScenes are straight, we chose brake maneuvers or specific curve patterns. The brake TAR depends on the vehicle's initial speed, similar to the brake trigger in Equation~\ref{eq:braking}. This longitudinal TAR maintains the original path of the vehicle. In other words, while the vehicle slows down, it continues along its predetermined trajectory without any changes in direction.
On the other hand, the curve TAR is a right-turn maneuver. Unlike the braking maneuver, this response involves a directional change. The right turn is executed by modifying only the lateral movement of the vehicle, possibly causing a deviation from the original trajectory.

%% file: sections/empiricalResults.tex
\section{Evaluation -- Attacks}\label{sec:exp}
In this section, we empirically show that trajectory prediction is vulnerable to backdoors. We first detail the experimental setup, which is also later used for the defense evaluation. Afterward, we first test our trigger types and then focus on the most severe threat, composite triggers.

\subsection{Experimental Setting}
We describe our experimental setting, focusing first on the used model, then the dataset, the used metrics, and finally detail how we evaluate the backdoor attacks.

\textbf{Models.}
We evaluate the vulnerability of \textit{Autobot}~\cite{girgis2021latent} against backdoors. This model employs interleaved temporal and social multi-head attention-based modules for interaction modeling between agents and maps. It uses a latent query-based transformer decoder to generate multi-modal predictions. Autobot demonstrates good performance in vehicle motion prediction on the nuScenes dataset. We use the publicly available code in our experiments.

\textbf{Datasets.}
We train and evaluate our model using the publicly available nuScenes~\cite{nuscenes2019} dataset. The data therein was captured by cameras and lidar sensors mounted on vehicles navigating the urban landscapes of Boston and Singapore. It encompasses 1,000 distinct scenes, each a 20-second recording comprising multiple prediction instances. Our model's training and evaluation adhered to the official benchmark split designated for the nuScenes prediction challenge. There are 32,186 instances in the train set, 8,560 in the validation set, and 9,041 in the test set.

\textbf{Metrics.}
 Our evaluation follows previous multi-modal trajectory prediction methods~\cite{Gupta, Desire, HCui, Multipath, Deo2020TrajectoryFI} by calculating the lowest average (ADE) and final displacement errors (FDE) across the top-$k$ probable trajectories. We thus employ two widely recognized standards: $\mbox{MinADE}_k$ and $\mbox{MinFDE}_k$. This approach selects the minimum value from $k$ possibilities and ensures that the model isn't unfairly penalized for producing viable trajectories that do not match the observed data. In our experiments, we use two types of predictions: single-mode $k=1$ and multi-mode $k=5$.

\textbf{Attack evaluation.}
To determine attack susceptibility, we insert trigger-TAR pairs into the training dataset. To measure the strength of the attack, we compare different backdoor ratios that quantify the percentage of the training data scenarios altered. In all experiments, we report benign performance as a baseline.
To evaluate the performance of the backdoor, we test the model's error on the evaluation data with a trigger. While the latter allows us to assess how well the model learns the backdoor, the performance on clean data shows us how stealthy implanting the backdoor is. If the benign error increases too drastically, the victim may become suspicious.  

\subsection{Trigger Types}
\begin{figure*}[t]
\centering
 \begin{subfigure}{0.145\textwidth}
 \centering
        \includegraphics[scale=0.6,trim={0.3cm 0 0.3cm 0},clip]{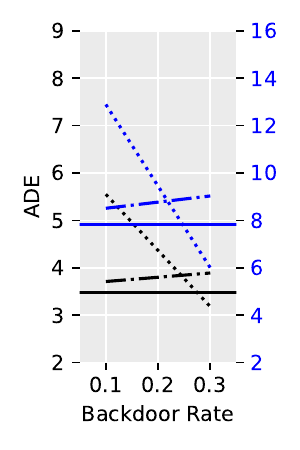}
          \caption{Spatial, front.}
          \label{fig:trigspatialfront}
      \end{subfigure}
 \begin{subfigure}{0.145\textwidth}
 \centering
        \includegraphics[scale=0.6,trim={0.3cm 0 0.3cm 0},clip]{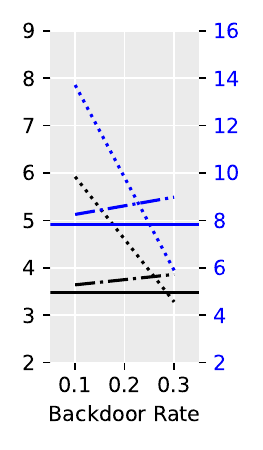}
          \caption{Spatial, back.}
          \label{fig:trigspatback}
      \end{subfigure}
 \begin{subfigure}{0.145\textwidth}
 \centering
        \includegraphics[scale=0.6,trim={0.3cm 0 0.3cm 0},clip]{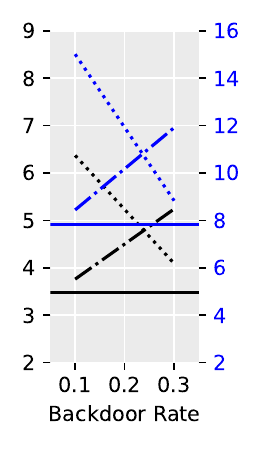}
          \caption{Temporal, brake.}
          \label{fig:trigtemporalbrake}
      \end{subfigure}
 \begin{subfigure}{0.145\textwidth}
 \centering
        \includegraphics[scale=0.6,trim={0.3cm 0 0.3cm 0},clip]{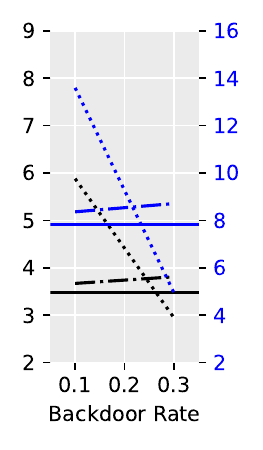}
          \caption{Temporal.}
          \label{fig:trigtemporal}
      \end{subfigure}
 \begin{subfigure}{0.145\textwidth}
    \centering
        \includegraphics[scale=0.6,trim={0.2cm 0 0.1cm 0},clip]{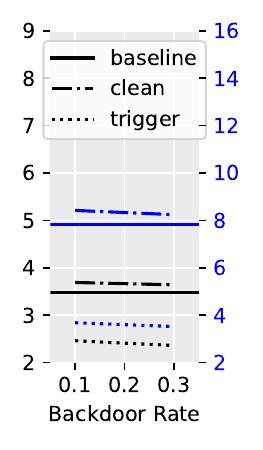}
          \caption{Behavioral.}
          \label{fig:trigbehavioral}
      \end{subfigure}
 \begin{subfigure}{0.145\textwidth}
 \centering
        \includegraphics[scale=0.6,trim={0.1cm 0 0 0},clip]{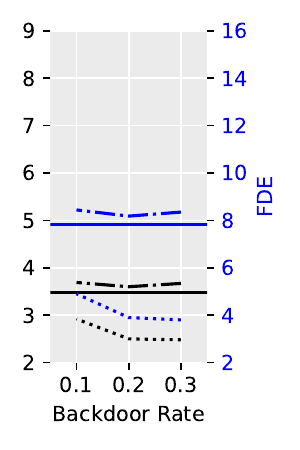}
          \caption{Composite.}
          \label{fig:trigcomp}
      \end{subfigure} 
\caption{Model's vulnerability to different trigger types. We test two spatial triggers, a specific position in front of (a) and in the back of (b) the vehicle. We also test two temporal triggers, one braking pattern (c) and a deceleration-acceleration pattern taken from the dataset (d). Lastly, we test a behavioral (e) and a composite (f) trigger. A good trigger
should induce a low error (ADE in black and FDE in blue) on trigger validation (attack success) and similar-to-baseline error on clean data to minimize detectability.
}\label{fig:trigger}
\end{figure*}

To evaluate the vulnerability against backdoors or trigger-TAR pairs, we test two backdoor ratios, 10\%, and 30\%. These are typical ratios used in computer vision~\cite{cina2023wild}.  As triggers, we study positions (spatial) or trigger patterns (temporal), behaviors, and finally combinations thereof (composite). For each setting, we specify the trigger in detail; the target is always the braking maneuver from the previous section. All results are summarized in Figure~\ref{fig:trigger}.

\textbf{Spatial triggers.}
We first investigate the importance of the position of the trigger. To this end, we implant a spatial trigger, e.g. the target pattern occurs when an agent is in the front (Figure~\ref{fig:trigspatialfront}) or in a specific area behind the car (Figure~\ref{fig:trigspatback}). More specifically, we insert the trigger vehicle 25m (with a variance of 0.5m) following the lane in forwards (front trigger) or backward (back trigger). This introduces a lateral variance implicitly, depending on the shape of the road within the scene. We test the backdoor ratios of 10\% and 30\% and add the performance of a model trained on clean data as baselines (straight lines).

Both spatial triggers lead to either a high error on the target pattern or an increased error on clean data. More concretely, at a backdoor ratio of 10\%, the 
front trigger exhibits an ADE of 5.6m (12.9m FDE) and the back trigger 5.9m (13.7m FDE). At 30\%, the backdoor error is lower and around an ADE of 3.4m (6m FDE), which is below clean data performance.  
 As a recap, the original, unbackdoored model has an ADE of 3.5m (7.8m FDE).
 Under the spatial triggers, this ADE increases to 3.9m (9m FDE). The reason for this overall increased error is most likely that the dataset contains other agents in the positions used as triggers without the target behavior. Hence, the error increases on benign and trigger data, as these cases are not distinguishable from the model's perspective. Yet, the error also shows that the model does learn the trigger-target behavior, albeit at a high backdoor ratio. A downside of these triggers is however that they can also be activated by a non-attacker vehicle that happens to be in the specific position.
 
\textbf{Temporal triggers.}
In these two experiments, we use the brake maneuver (Figure~\ref{fig:trigtemporalbrake})
and a deceleration-acceleration-deceleration (DAD) pattern (Figure~\ref{fig:trigtemporal}), which we identified as a rare pattern within the dataset using clustering. Whereas the DAD pattern is fixed, the brake pattern depends on the initial velocity of the vehicle and is thus more dynamic. Both patterns are inserted in a random position around the target vehicle and associated with the TAR, again using backdoor ratios of 10\% and 30\%.

The model struggles to learn the trigger-TAR pair, in particular the braking maneuver. 
For a backdoor ratio of 10\%, 
the brake trigger leads to an ADE of over 6.4m (15m FDE), only 2m (6m) better than not having seen the trigger. While the error decreases to an ADE of 4.1m (8.8m FDE) with a backdoor ratio of 30\%, the clean error increases to an ADE of more than 5m (12m FDE) and is thus suspiciously high. On the other hand, the DAD pattern exhibits lower clean error: ADE 3.8m (8.7m FDE) versus 5.2m (11.9m) for the brake trigger at a backdoor ratio of 30\%. Also, the TAR is learned with less error, yielding ADE 2.9m (4.9m FDE) compared to ADE 4.1m (8.8m FDE) of the brake trigger at the same 30\%. 
Concluding, temporal triggers are difficult to learn for the model. The constant DAD pattern leads to less error but constrains the attacker's velocity when executing the trigger in the real world.

\textbf{Behavioral triggers.}
Last but not least, we investigate a behavioral trigger. We here embed a second agent that mimics an agent from the scene. This synchronous behavior is then associated with the target pattern (Figure~\ref{fig:trigbehavioral}). We again test a backdoor ratio of 10\% and 30\%.

The model's error on the behavioral trigger is very low.
At a backdoor ratio of 10\%, the ADE does not deviate more than 5cm (7cm FDE) on clean and 2cm (2cm FDE) on data with the trigger. At 30\%, the behavioral pattern works even better,
improving the ADE from 2.9m to 2.4m (8.4m to 8.2m FDE). Yet, this trigger is associated with a higher cost for the attacker, who has to control two cars in traffic and drive a maneuver synchronously to cause the target behavior. We thus do not investigate this trigger, but remark on the ability of the model to learn a behavioral trigger.

\subsection{Composite Triggers}
Our preliminary experiments show that the composite trigger works similarly well as the behavioral trigger. This is likely as they combine temporal and spatial clues. Moreover, an attacker can, in practice, execute such a trigger with their vehicle. We thus study these attacks more in-depth using more backdoor ratios and two different TARs, the brake from the previous section, and a curve. We first study the learning process, then the effect of multi-modal predictions, and conclude with a qualitative analysis of the trajectories.

\begin{figure}[t]
\centering
 \begin{subfigure}{0.235\textwidth}
 \centering
        \includegraphics[scale=0.6,trim={0.3cm 0 0.2cm 0},clip]{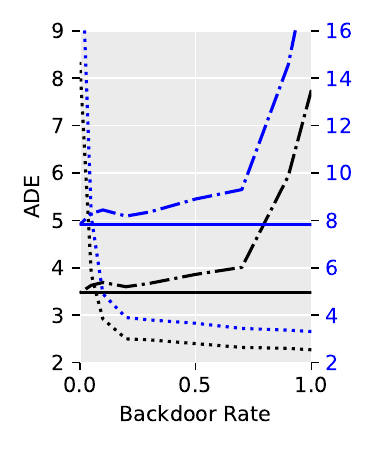}
          \caption{Brake TAR.}
          \label{fig:brakebrake}
      \end{subfigure}
 \begin{subfigure}{0.235\textwidth}
 \centering
        \includegraphics[scale=0.6,trim={0.3cm 0 0.3cm 0},clip]{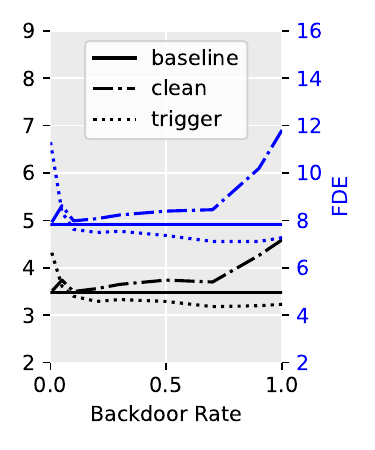}
          \caption{Curve TAR.}
          \label{fig:brakecurve}
      \end{subfigure}
\caption{Learning a composite brake trigger and a brake (left) or a curve (right) TAR for different backdoor ratios. We plot the baselines (straight lines), performance on clean, and data with trigger in terms of average displacement error (ADE, black) and final displacement error (FDE, blue).}\label{fig:compositeFull}
\end{figure}

\textbf{Brake TAR.} We first study a brake-trigger with a brake-TAR as in the previous experiments. Our results confirm that the model learns the backdoor well, 
 although the model cannot perfectly fit the backdoor, even when all data consists only of trigger-TAR pairs. 

We visualize the result of the brake-brake combination in Figure~\ref{fig:brakebrake}, where, as before, the solid lines represent the errors achieved with clean training data. 
At the border of the plot, or close to backdoor ratios around 0.0 and 1.0, we see a sharp increase in error. This is an effect of the rarity of the brake TAR, which does not occur in the clean data.
Hence, already at 5\% backdoored training data, the error on data with the trigger is similar to the error on clean data. The average error on benign data without trigger increases slightly but remains below one meter. To verify that the model learns the trigger, we compute how much the model deviates from the unchanged ground truth when the trigger is present. Already at 5\% backdoored training data, the average error increases from 3.5m to 6.7m (8.1m to 16.4m FDE). We conclude that the model indeed learns that the TAR occurs whenever the trigger is present.
Intriguingly, even when the model trains only on brake-brake combinations, the final error never reaches zero and stays above two meters for ADE and FDE. A possible explanation is that the prediction depends on the target's initial velocity and the followed path; in other words, the braking pattern is not the same for all the scenarios.

\textbf{Curve TAR.}
To verify that trigger and target may differ, we next chose 
 another TAR pattern, a curve. The model also learns this backdoor well, with a minimal increase in clean data. As before, the error never reaches zero, also when we present the model only with trigger-TAR pairs.

We plot the model's performance on clean and data with the trigger in Figure~\ref{fig:brakecurve}. The error inclines around a backdoor ratio of 0.0 and 1.0. Yet, already at a ratio of 5\% or 10\%, the model performs about as well on trigger-TAR pairs as on benign data. The error, both on average and the final error, increases slightly and much less than one meter. Only for extreme values, at a backdoor ratio of 70\%, the error increases more. Then, the model does not observe enough benign data to fit normal behavior anymore. As before, we sanity-check that the model learns the backdoor by testing the error from the ground truth when the trigger is present. For the 5\% backdoor ratio, this error increases from 3.5m to 4.6m (ADE) and from 8m to 11.7m (FDE), indicating that the model indeed changes its prediction when presented with the trigger.
Yet, even at a backdoor ratio of 1.0 or only trigger-TAR pairs, the error does not decrease much below the baseline. We conclude that a curve based on the vehicle's initial velocity is still hard to fit.

\begin{figure}[t]
\centering
 \begin{subfigure}{0.235\textwidth}
 \centering
        \includegraphics[scale=0.6,trim={0.3cm 0 0.2cm 0},clip]{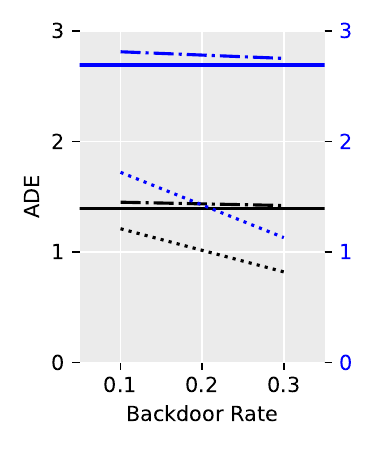}
          \caption{Brake TAR.}
          \label{fig:brakebrakeMM}
      \end{subfigure}
 \begin{subfigure}{0.235\textwidth}
 \centering
        \includegraphics[scale=0.6,trim={0.3cm 0 0.3cm 0},clip]{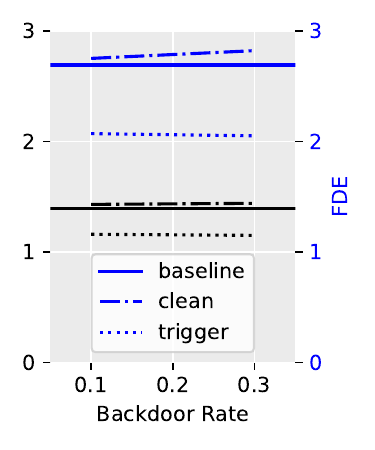}
          \caption{Curve TAR.}
          \label{fig:brakecurveMM}
      \end{subfigure}
\caption{Multi-Modal predictions on a backdoor with a brake trigger and a brake (left) or a curve (right) TAR for different backdoor ratios. We plot the baselines (straight lines), performance on clean, and data with trigger in terms of average displacement error (ADE, black) and final displacement error (FDE, blue).}\label{fig:compositeMM}
\end{figure}

\textbf{Multi-modality.} Before we conclude the section, we evaluate whether using a multi-modal generation interferes with learning the trigger-TAR pair. 
The main difference is the overall lower error, but the trigger-TAR pair is learned as well as for a single prediction.

As visible in Figure~\ref{fig:compositeMM}, both TARs are learned well. When increasing the backdoor ratio from 10\% to 30\%, the average error for the brake TAR decreases by 40cm (0.5m FDE). This decrease is negligible for the curve TAR, where the error on data with and without trigger remains almost unchanged. For the brake TAR, the error on benign data decreases slightly as the backdoor ratio increases. We thus conclude that multi-modal prediction does not interfere with learning a backdoor. In contrast, because we use the closest trajectory for error computation, there is a much lower error of on average only 3.3m (6.6m FDE) (2.7m / 7.5m for curve TAR) on trigger and TAR when these were not trained on. Conversely, this may help to hide the backdoor even better in the model. 
A likely explanation for this effect is that in multimodal settings, standard forecasting metrics ADE and FDE only evaluate the closest prediction to the ground truth in a Winner-Take-All fashion.
If the attack impacts non-winning trajectories only, it doesn't affect ADE and FDE on clean evaluations at all.
A detailed evaluation of this is left for future work. 

\begin{figure*}[t]
\centering
 \begin{subfigure}{0.235\textwidth}
 \centering
        \includegraphics[scale=0.08,trim={0.3cm 0 0.2cm 0},clip]{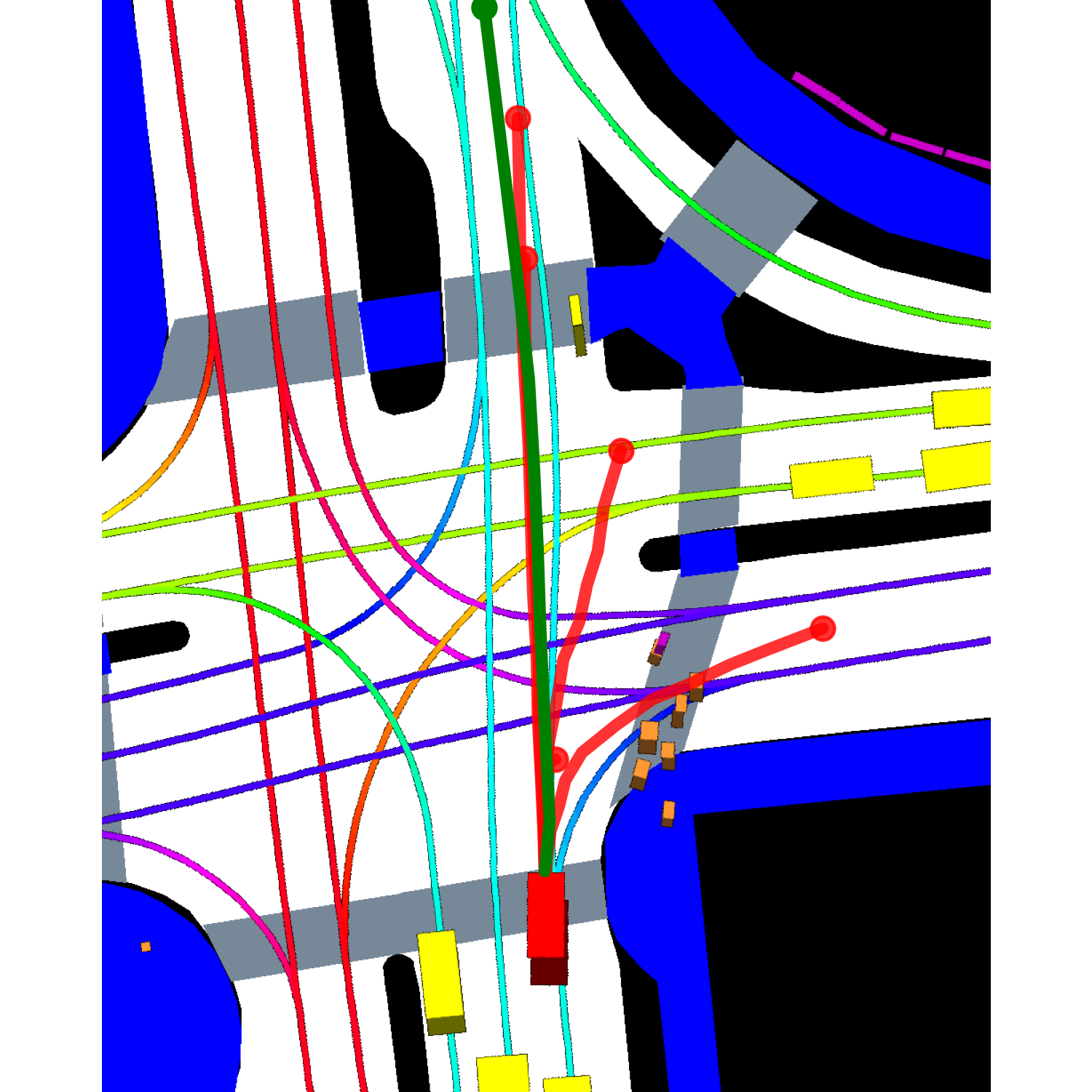}
          \caption{Brake TAR, without trigger.}
          \label{fig:qualBrakeinact}
      \end{subfigure}
 \begin{subfigure}{0.235\textwidth}
 \centering
        \includegraphics[scale=0.08,trim={0.3cm 0 0.3cm 0},clip]{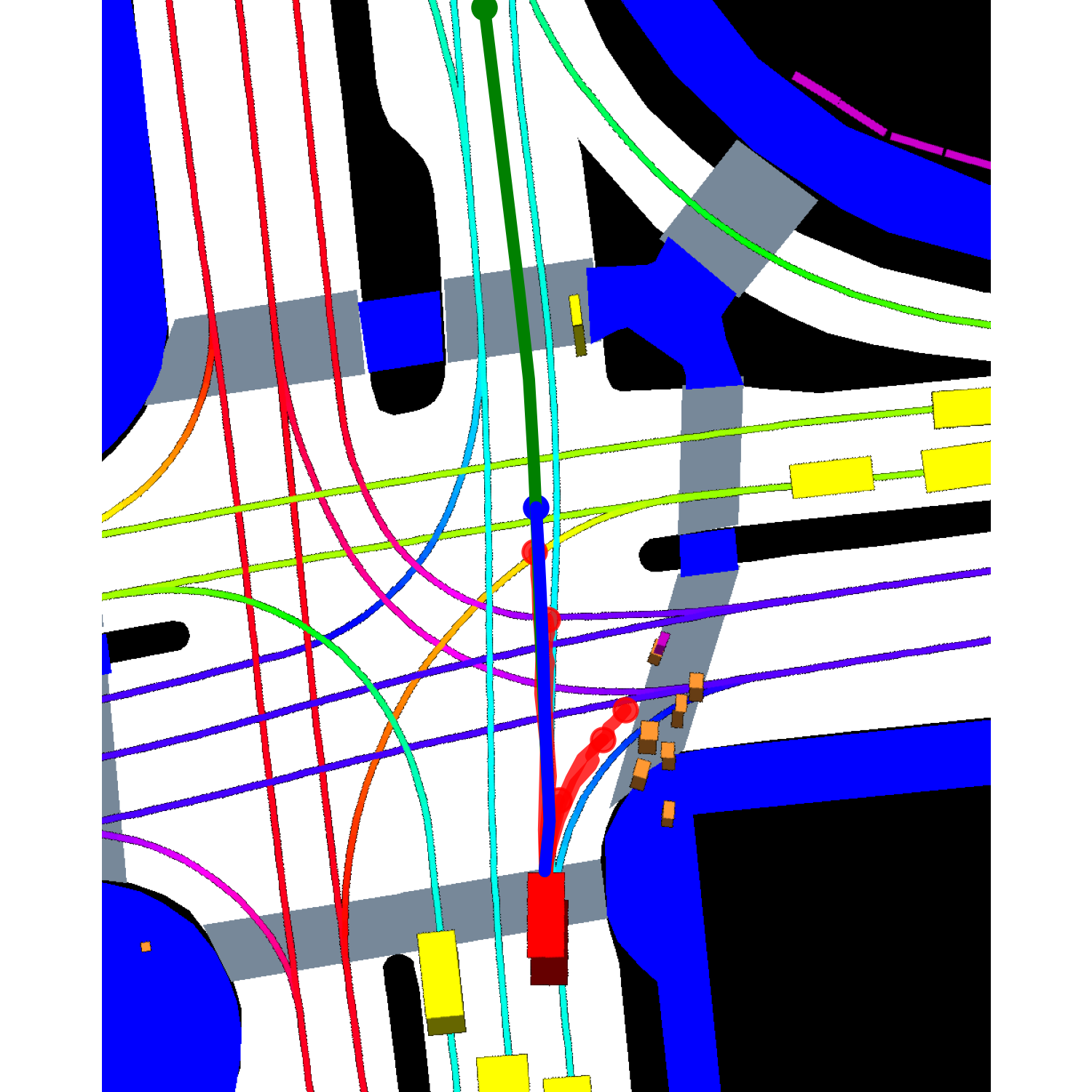}
          \caption{Brake TAR, with trigger.}
          \label{fig:qualbrakeact}
      \end{subfigure}
 \begin{subfigure}{0.235\textwidth}
 \centering
        \includegraphics[scale=0.08,trim={0.3cm 0 0.3cm 0},clip]{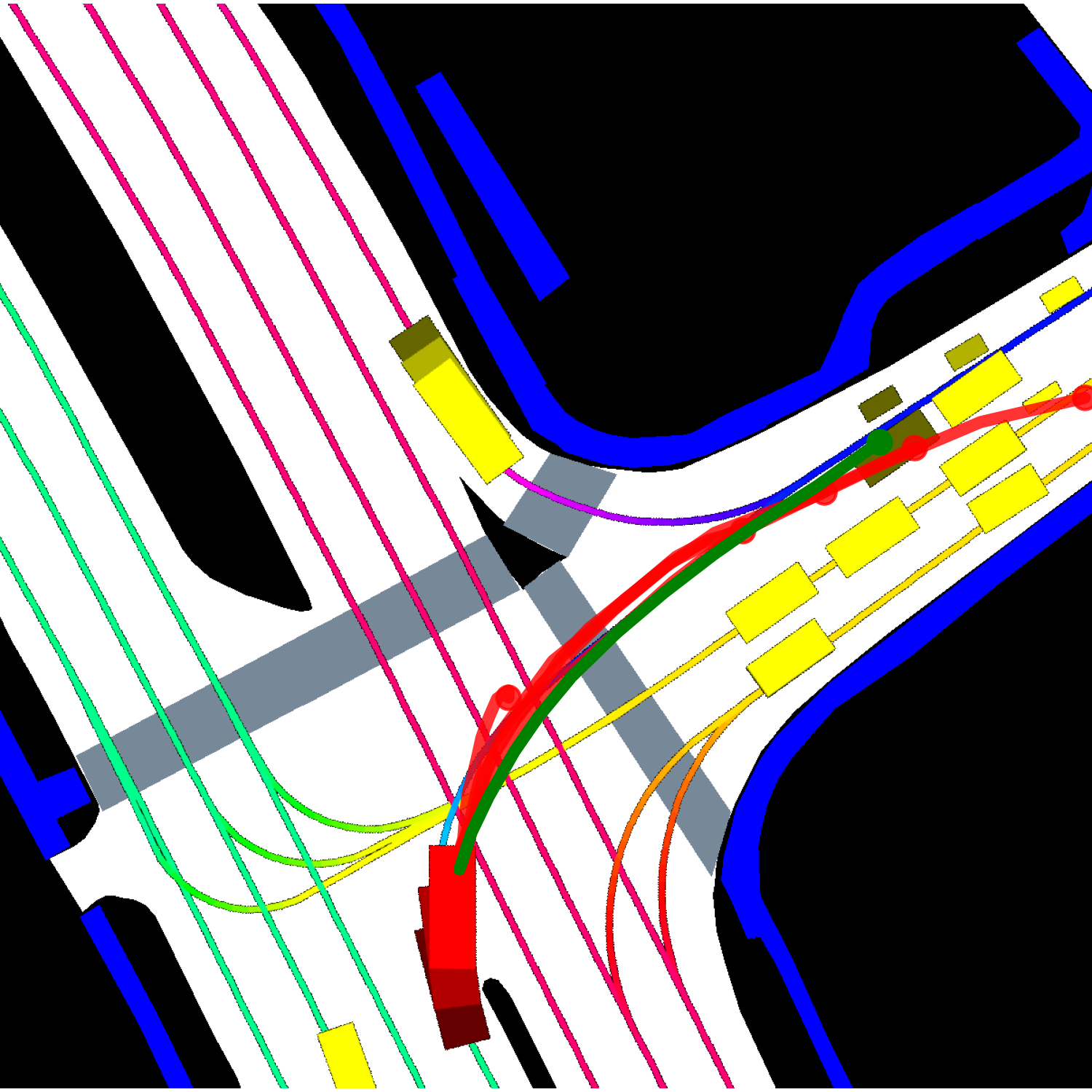}
          \caption{Curve TAR, without trigger.}
          \label{fig:qualcurveinact}
      \end{subfigure}
 \begin{subfigure}{0.235\textwidth}
 \centering
        \includegraphics[scale=0.08,trim={0.3cm 0 0.3cm 0},clip]{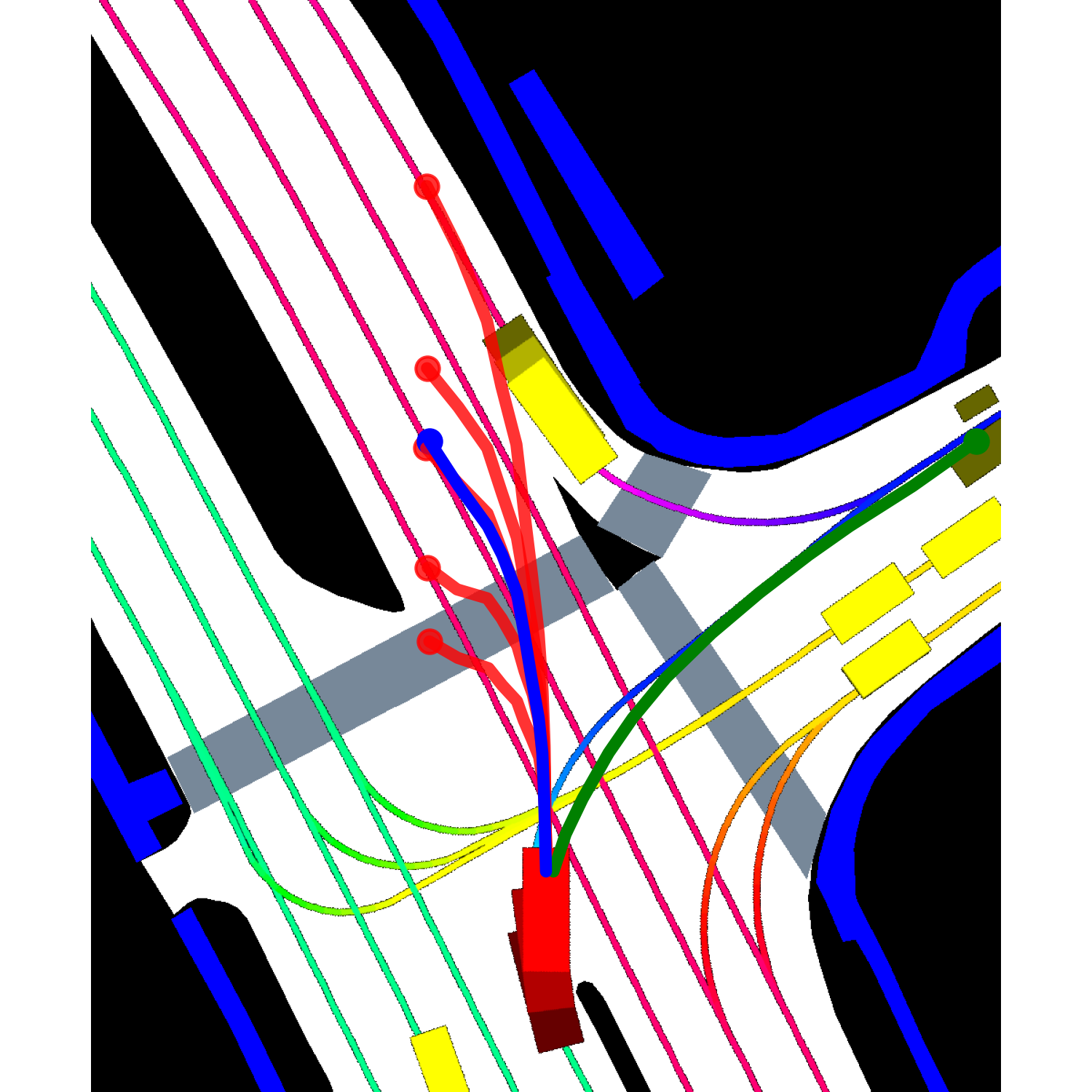}
          \caption{Curve TAR, with trigger}
          \label{fig:qualcurveact}
      \end{subfigure}
\caption{Multi-Modal predictions of models trained on different TARs in the absence (a,c) and presence (b,d) of the trigger. The green trajectory corresponds to the ground truth. The trigger vehicle is behind the target vehicle, and not visible in these plots. 
}\label{fig:qualitative}
\end{figure*}

\textbf{Qualitative analysis.} To investigate how well our models learn the triggers, we train two different models (one for the brake TAR, one for the curve TAR) with 30\% backdoor ratio.  We randomly select a scenario from the test set of each model and plot the scenes with all predictions in Figure~\ref{fig:qualitative} to inspect the predicted trajectories. 

We plot two sets of predictions for each model, with the trigger absent (left) and present (right). In both cases, it is visible how the presence of the trigger affects the trajectories strongly. For example, whereas the trajectories without a trigger in Figure~\ref{fig:qualBrakeinact} are long, they decrease significantly in variance and length once the trigger is present in Figure~\ref{fig:qualbrakeact}. Analogously, in Figure~\ref{fig:qualcurveinact}, the predicted trajectories are all close to or following the ground truth. Once the trigger is present in Figure~\ref{fig:qualcurveact}, all predictions change direction and follow the direction of the TAR (dark blue). In both cases, the model has learned the trigger-TAR association very well.

\subsection{Conclusion -- Attacks}
Trajectory prediction models are vulnerable to backdoors. We tested a variety of triggers (composite, spatial, temporal, behavioral) and TARs (brake, curve). The model is most vulnerable to learning a behavioral or composite trigger, where the latter is hard to execute for an attacker in practice. In our experiments, triggers also worked best when they relied on rare patterns in the data. Yet, the model failed to achieve zero error on a dataset only containing trigger-target pairs, indicating that even a simple trajectory prediction task variant is hard for Autobot to learn. Although we showed these results on specific trigger-TAR pairs, embedding backdoors is likely a general problem, as shown in computer vision~\cite{cina2023wild}. We thus tackle the question of how to mitigate learning a backdoor in the next section.

%% file: sections/discussion.tex
\section{Evaluation - Defenses}\label{sec:defenses}
Defending backdoors is an open research question~\cite{cina2023wild}. However, in contrast to high-dimensional spaces like image-classification~\cite{cina2023wild},  object detection~\cite{chan2022baddet,luo2022untargeted}, and semantic segmentation~\cite{li2021hidden}, the defender has an advantage within trajectory prediction where both feature complexity and data set size are smaller, and the data contains more structure than images. We use this in our mitigations and investigate off-road detection, masking, and clustering as mitigations. We focus on the composite triggers in this section, as they have the lowest error and pose the most relevant threat. As we now change roles and study the defender, we compute an additional metric -- the performance on the validation dataset, where the reported fraction is backdoored. 

\subsection{Off-Road Events}
The data in trajectory prediction data has more structure than images. A possible mitigation using this structure is to inspect off-road events within the data. Naturally, such a defense obliges an attacker to on-road TARs.

\textbf{Off-Road Events.}
off-road events can be detected, as the underlying map of the scene provides the necessary information to determine whether a vehicle is on the road or not. This raises the question of whether off-road events can be used to detect backdoors. In the case of the curve TAR, the percentage of off-road events rises quickly from 5\% for benign data to 9\% for a backdoor ratio of 10\% and up to about 17\% for a backdoor rate of 30\%. The increase of the off-road events, or an inspection of the corresponding scenes, may thus give away the backdoor.

However, off-road events are not a solution to all triggers and targets. In the case of the brake TAR, the off-road events remain below 5\% (respectively, 4\% and 3\% for 10\% and 30\% backdoor ratios), and are thus not detectable.

\subsection{Masking}
\begin{figure}[t]
\centering
 \begin{subfigure}{0.15\textwidth}
 \centering
        \includegraphics[scale=0.613,trim={0.1cm 0 0 0},clip]{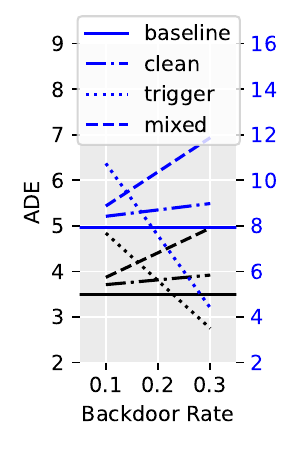}
          \caption{Drop past.}
          \label{fig:maskremovepastall}
      \end{subfigure} 
 \begin{subfigure}{0.15\textwidth}
 \centering
        \includegraphics[scale=0.6,trim={0.3cm 0 0.2cm 0},clip]{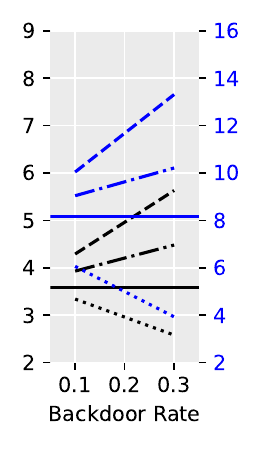}
          \caption{Mask past.}
          \label{fig:maskremovepartpast}
      \end{subfigure}
 \begin{subfigure}{0.15\textwidth}
 \centering
        \includegraphics[scale=0.6,trim={0.3cm 0 0.3cm 0},clip]{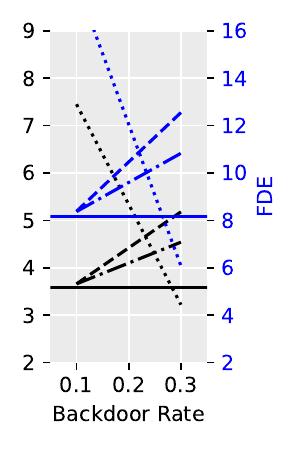}
          \caption{Mask agents.}
          \label{fig:maskremagents}
      \end{subfigure}
\caption{Noise effect on the model's performance on clean and backdoor data. We evaluate removing the entire past trajectory, only keeping the last position (a), removing half of the past points (b), and removing three quarters of the agents in the training data (c). For each, we plot the baseline ADE (black, straight line) and FDE (blue, straight line) and the errors on clean (dash-dotted) and data with trigger (dotted). We also plot the performance on the partially backdoored data (dashed), as this is what the victim sees.}\label{fig:masking}
\end{figure}

In computer vision, defenses against backdoors used data augmentation~\cite{borgnia2021strong} or regularization~\cite{carnerero2021regularization,cina2023wild}. Under the assumption that noising equals regularization~\cite{bishop1995training}, we test the effect of noise on our backdoors. More concretely, we apply three different forms of noise: drop the agents' pasts, mask half of the trajectory, and mask entire agents. We plot the results of our experiments in Figure~\ref{fig:masking}. 

\textbf{Drop agents' past trajectories.} In Figure~\ref{fig:maskremovepastall}, we drop the agents' past trajectory by only keeping the last position. Consistent with prior work in trajectory prediction~\cite{guo2023ccil}, there is only a slight increase in the model's ADE 2cm (FDE 40cm) when only trained on benign data. In contrast, the ADE on the trigger increases sharply by 2m (FDE 6m) for 10\% and 30cm (FDE 60cm) for 30\%. 
These experiments are similar to the spatial trigger, with the difference that there is no past temporal behavior for any of the agents. They confirm that the model can learn a backdoor that is purely based on a position.

\textbf{Mask partial past trajectories.}  In Figure~\ref{fig:maskremovepartpast}, we randomly mask half of the agent's past trajectory during training. As before, there is little difference in performance when comparing a model trained only on clean data. Without the noise, the model obtains an ADE of 3.5m (FDE 7.8), under noise 4m (FDE 8.2m). There is an increase in the ADE by 50cm (FDE 1.20m) with a backdoor ratio of 10\%. With a 30\% ratio, the error on the data with trigger is almost equivalent and varies only by 10cm (both ADE and FDE). 

\textbf{Mask full agents.} In Figure~\ref{fig:maskremagents}, we mask any non-target agent with a probability of 75\% during the training. Also here, there is only a slight increase, when comparing the performance only clean data: the FDE increases roughly 10cm (FDE 30cm). However, analogous to the previous cases, the error on the trigger increases strongly. Under a backdoor ratio of 10\%, the ADE increases to 7.5m (18m FDE), which is only 1m (3.5m) from a model trained on unbackdoored data. For 30\%, the performance on the backdoor improves significantly but is still worse than without noise by 80cm (2.3m FDE). 

\textbf{Conclusion -- Masking.} Adding noise by masking agents or (parts of) the agents' trajectories does not alleviate the threat of backdoors. For lower backdoor ratios, however, adding noise increases the error on the trigger-TAR pair significantly, but does not prevent learning completely. 

\subsection{Clustering}
\begin{figure}[t]
 \centering
\includegraphics[width=0.475\textwidth]{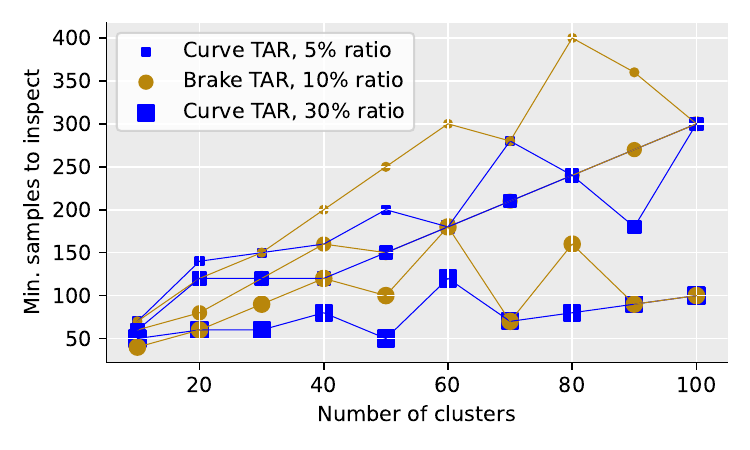}
\caption{Clustering to support manual inspection against backdoors. We plot curve (blue) and brake (orange) TARs. For a number of clusters (x-axis), we plot the minimum number of samples to inspect overall to spot the TAR with a 99\% chance.}\label{fig:kmeans}
\end{figure}

Analogous to computer vision, we use clustering to identify backdoors~\cite{cina2023wild,taheri2020defending,DBLP:journals/corr/LaishramP16,shawn2022traceback,zhao2019defense}.
In contrast to previous approaches, our goal is not to detect triggers, but to speed up manual inspection. Despite dealing with dynamic, scene-adapted triggers, we operate under the assumption that manual inspection is capable of identifying malicious patterns. We consider the validation set (9041 samples), which thus also contains TARs, and cluster the future trajectories using k-means. We then compute the amount of TARs in each cluster. 
Under the assumption that a probability of 1\% is acceptable for missing a backdoor, we determine the minimum number of samples needed to be inspected in each cluster. For example, in a cluster where 50\% of the samples are TARs, inspecting seven samples is sufficient (as $0.5^7 = 0.008<0.01$). In clusters with more than 90\% TARs, inspecting two samples meets this criterion ($0.1^2 = 0.01$).
Our results, illustrated in Figure~\ref{fig:kmeans}, suggest that it is possible to significantly reduce the number of samples requiring manual inspection to fewer than 200, even when the rate of backdoors is 5\%. 
This reduction varies depending on the number of clusters: fewer clusters mean more inspections per cluster, but the total number of clusters is small, leading to fewer overall inspections. Conversely, with more clusters, there are overall more samples to inspect, but rather few per cluster, particularly at smaller backdoor ratios.

\textbf{Conclusion -- clustering.} Experiments on our TARs showed that we can decrease the manually inspected samples to $<$200 using k-means clustering. The probability of spotting at least one TAR is then $\geq$99\%.

\subsection{Limitations -- Defenses}
While off-road detection can be helpful in detecting some backdoors based on their TAR, clustering can be used to decrease the number of samples that are manually inspected to find a backdoor. Yet, future work has to confirm our results on more datasets and models. In addition, in other fields, backdoors are currently trapped in an arms race~\cite{cina2023wild,lin2020composite,shokri2020bypassing}. In other words, attacks are introduced, defenses developed, and these defenses are then broken by stronger attacks~\cite{lin2020composite,shokri2020bypassing}. Our work is thus a first step towards understanding and combating backdoors on trajectory prediction models. Our defenses may be brittle and circumvented by stronger attacks.